%% file: neurips_2023.tex
\documentclass{article}

\usepackage{graphicx}
\usepackage{amsmath}
\usepackage{amssymb}
\usepackage{booktabs}
\usepackage{wrapfig}

\usepackage{multirow}
\usepackage{verbatim}
\usepackage{enumitem}
\usepackage{bm}
\usepackage{arydshln}
 \usepackage{float} 
\usepackage{subcaption}

\usepackage{xcolor}
\definecolor{bittersweet}{rgb}{1.0, 0.44, 0.37}
\definecolor{mygreen}{rgb}{0.29, 0.7, 0.48}

\newcommand{\tabincell}[2]{\begin{tabular}{@{}#1@{}}#2\end{tabular}}

\usepackage[font=small,labelfont=bf]{caption}  
\usepackage{makecell}
\usepackage{tabulary}
\definecolor{demphcolor}{RGB}{144,144,144}
\newcommand{\demph}[1]{\textcolor{demphcolor}{#1}}
\definecolor{mygray}{gray}{0.4}
\usepackage{pifont}
\newcommand{\cmark}{\color{mygray}\ding{51}}%
\newcommand{\xmark}{\color{mygray}\ding{55}}%

\usepackage{xspace}

\newcommand\blfootnote[1]{%
  \begingroup
  \renewcommand\thefootnote{}\footnote{#1}%
  \addtocounter{footnote}{-1}%
  \endgroup
}

\usepackage[pagebackref,breaklinks,colorlinks]{hyperref}

\usepackage[preprint]{neurips_2023}

\usepackage[utf8]{inputenc} 
\usepackage[T1]{fontenc}    
\usepackage{hyperref}       
\usepackage{url}            
\usepackage{booktabs}       
\usepackage{amsfonts}       
\usepackage{nicefrac}       
\usepackage{microtype}      
\usepackage{xcolor}         

\newcommand{\ie}{{i}.{e}.}
\newcommand{\eg}{{e}.{g}.}

\usepackage{enumitem}
\usepackage{xcolor}
\definecolor{blue}{rgb}{0,0,1}

\definecolor{red}{rgb}{1,0,0}

\definecolor{black}{rgb}{0,0,0}

\title{ConES: Concept Embedding Search for Parameter Efficient Tuning Large Vision Language Models}

\author{
Huahui Yi$^{1*}$\quad Ziyuan Qin$^{1*}$\quad Wei Xu$^{1,2}$ \quad Miaotian Guo$^{3}$\quad Kun Wang$^{4}$\\
\textbf{Shaoting Zhang}$^{6}$\quad \textbf{Kang Li}$^{1,5,6{\dagger}}$\quad \textbf{Qicheng Lao}$^{3,6{\dagger}}$\\
$^1$West China Biomedical Big Data Center, West China Hospital, Sichuan University\\ 
$^2$School of Biomedical Engineering \& Suzhou Institute for Advanced Research, USTC\\
$^3$School of Artificial Intelligence, BUPT~~
$^4$School of Data Science, USTC\\
$^5$Sichuan University Pittsburgh Institute~~~~~~~~~
$^6$Shanghai AI Laboratory\\
\texttt{qicheng.lao@bupt.edu.cn}
}

\begin{document}

\maketitle

\begin{abstract}
Large pre-trained vision-language models have shown great prominence in transferring pre-acquired knowledge to various domains and downstream tasks with appropriate prompting or tuning.
Existing prevalent tuning methods can be generally categorized into three genres: 1) prompt engineering by creating suitable prompt texts, which is time-consuming and requires domain expertise; 2) or simply fine-tuning the whole model, which is extremely inefficient; 3) prompt tuning through parameterized prompt embeddings  with the text encoder. Nevertheless, all methods rely on the text encoder for bridging the modality gap between vision and language. In this work, we question the necessity of the cumbersome text encoder for a more lightweight and efficient tuning paradigm as well as more representative prompt embeddings closer to the image representations. To achieve this, we propose a \textbf{Con}cept \textbf{E}mbedding \textbf{S}earch (\textbf{ConES}) approach by optimizing prompt embeddings---without the need of the text encoder---to capture the `concept' of the image modality through a variety of task objectives. By dropping the text encoder, we are able to significantly speed up the learning process, \eg, from about an hour to just ten minutes in our experiments for personalized text-to-image generation without impairing the generation quality. Moreover, our proposed approach is orthogonal to current existing tuning methods since the searched concept embeddings can be further utilized in the next stage of fine-tuning the pre-trained large models for boosting performance. Extensive experiments show that our approach can beat the prompt tuning and textual inversion methods in a variety of downstream tasks including objection detection, instance segmentation, and image generation. Our approach also shows better generalization capability for unseen concepts in specialized domains, such as the medical domain.\blfootnote{$^*$Equal contribution. ~~~~~~~$^\dagger$Corresponding author.}

\end{abstract}

\section{Introduction}
Vision-Language Models (VLMs) have emerged as powerful models for bridging the gap between visual and textual input, witnessed significant advancements in terms of their generalization capability and the transferability of their learned representations~\cite{clip,align,glipv1,glipv2,uninext}. By capturing high-level semantic concepts shared across different modalities, VLMs can transfer their learned knowledge to various domains with transferable text prompts encapsulating high-level semantics that align with visual concepts. 
Therefore, the quality of the text prompts through prompt learning is crucial for the VLM's performance on many visual tasks.  
However, finding suitable prompts is a non-trivial task, and a slight change in word selection in the prompts can lead to a dramatic difference in performance. 

\begin{figure*}
  \centering
    \includegraphics[width=1.0\linewidth]{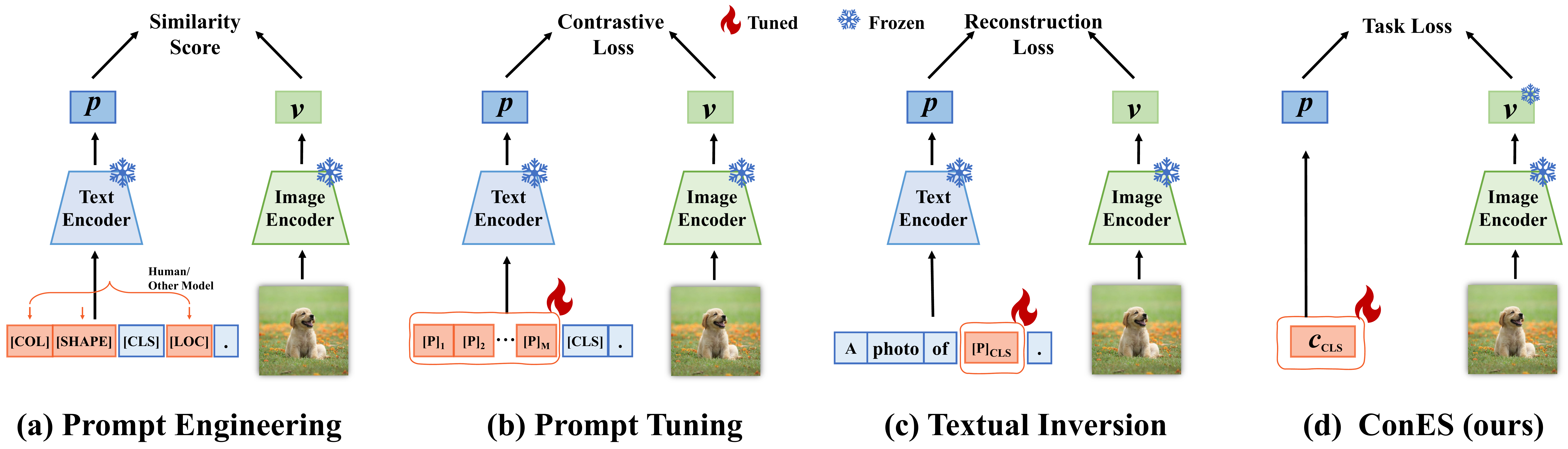}
  \caption{Graphical illustration of existing methods (a-c) and our proposed method (d), concept embedding search, or ConES for short. In contrast to other approaches, ConES does not use Text Encoder.}
  \label{fig:4methods}
  \vspace{-6mm}
\end{figure*}

The most apparent method of finding appropriate prompts for VLMs is prompt engineering (Figure~\ref{fig:4methods} (a)), and the idea is straightforward by simply making changes explicitly on the prompt texts~\cite{clip,shin2020autoprompt,reynolds2021prompt,llmprompt}. This process can be seen as word tuning and is extremely time-consuming and often requires expertise in the target domain to design the prompts, such as the medical domain~\cite{iclr2023,MEDIMPMI,Sivarajkumar2022HealthPromptAZ}. 
Another school of thought in tuning large VLMs is soft prompt tuning~\cite{coop,cocoop,fiber,vpt}, which focuses on integrating parameterized token embeddings with the given prompts and tuning the embeddings (Figure~\ref{fig:4methods} (b)). These approaches have breakthroughs in the parameter efficiency of tuning VLMs since a large part of the model is untouched while only prompt embeddings are tuned. 
Though there are many variants that put the parameterized tokens at different positions of the prompt or different layers of the text encoder~\cite{coop,prompt-tuning,p-tuning}, they all rely on a large text encoder to optimize the prompt embeddings. 

The essential mechanism of prompt tuning methods is to train a set of learnable embeddings through a large pre-trained text encoder, 
making learned prompt embeddings share the same embedding space with the language encoder.  to question the necessity of learning embeddings through text encoder. 
However, as mentioned in~\cite{mindgap}, due to the cone effect and the temperature hyper-parameter in contrastive loss, there exists a modality gap between the text and image encoders where the VLM's performance is shown to be sensitive to the change of modality gap~\cite{mindgap}. 
Therefore, we argue that the modality gap problem is inevitable in current prompt tuning methods due to the adoption of the text encoder for obtaining the prompt embeddings.
This phenomenon also intrigues us to question the necessity of learning embeddings through a large pre-trained text encoder, and whether it is possible to bypass the cumbersome text encoder to obtain a prompt embedding directly.

To answer this question, in this work, we propose a more lightweight and efficient parameter tuning paradigm for VLMs, which we call the Concept Embedding Search (ConES) approach (Figure~\ref{fig:4methods}~(d)), for learning more representative prompt embeddings closer to visual concepts without the text encoder.
Concretely, our proposed approach initializes a fixed number of random parameterized tokens which is the same shape as the standard prompt token embeddings. Given the image input and a frozen image encoder from the pre-trained VLMs, these tokens are optimized to learn the visual concepts captured in the image modality through a wide range of objective losses depending on the task, \eg, reconstruction loss for the image generation task.
After embedding search in the first stage, we obtain a set of token embeddings that is close to the visual concept representation, \ie, concept embeddings, and then we can carry these concept embeddings to downstream tasks to further fine-tune the VLMs without text encoder for boosting performance. 
Overall, this paper makes the following contributions:
\begin{itemize}[leftmargin=*]
    \item To the best of our knowledge, our approach is the first parameter-efficient fine-tuning (PEFT) method that completely bypasses the text encoder to search for effective prompt embeddings. Removing the text encoder decreases the number of parameters significantly during the forward pass process, making our method the most efficient method.
    \item  We demonstrate that the searched concept embeddings can mitigate the modality gap impact and show superior performance on various tasks compared to other PEFT methods. We also demonstrate with experiments that our method has better generalization capability for unseen concepts.
    \item 
    We conduct extensive experiments on 24 datasets for three different tasks including object detection, instance segmentation, and personalized text-to-image generation to show that our approach can achieve remarkable results while halving the parameter size of the VLMs.
\end{itemize}

\section{Related Work}
\noindent\textbf{Vision-Language Models}
Vision-Language Models (VLMs) have gained much popularity in recent years as a significant improvement in cross-modality artificial intelligence. Some notable pioneer works, such as CLIP~\cite{clip} and ALIGN~\cite{align}, focus on aligning visual concepts with human language by leveraging the web-scale of paired images and their text description. Compared to traditional visual models, who are trained with closed-set class labels, VLMs like CLIP~\cite{clip} perform well on unseen or out-of-distribution data. This ability is largely attributed to the rich semantic information contained in the human language. Furthermore, GLIP~\cite{glipv1} is one of the first batches of VLM works designed for solving object detection and grounding tasks with the help of text prompts. This work reformulates the object detection task as phrase grounding by optimizing the alignment scores between regions and words in the prompt. UNINEXT~\cite{uninext} further proposes to reformulate instance perception tasks into a unified object discovery and retrieval paradigm. In this work, we use the GLIP and UNINEXT models as base models to test the effectiveness and efficiency of our method.

\noindent\textbf{Prompt Tuning} Prompt tuning is an efficient and flexible approach for adapting pre-trained models to specific tasks and domains with minimal additional parameters and data, achieved by adding tunable prefix tokens to the input or hidden layers and training only these soft prompts when fine-tuning on downstream tasks. This method gained popularity in the NLP domain, leading to notable works such as pre-fix tuning~\cite{prefix_tuning}, prompt tuning~\cite{prompt-tuning}, and p-tuning~\cite{p-tuning}, and has also shown promise in the computer vision and vision-language domain with works like VPT~\cite{vpt}, S-Prompting~\cite{spt}, CoOp~\cite{coop}, and CoCoOp~\cite{cocoop}.

\noindent\textbf{Textual Inversion} Textual Inversion~\cite{text_inversion} captures novel concepts from a small number of example images by learning new `word' in the text encoder's embedding space, enriching personalized image generation. This approach can synthesize new concepts and attributes not in the training data, capturing higher-level semantic meaning. Validation on multiple variants of the diffusion model demonstrates its effectiveness in enhancing control over images generated from text-to-image pipelines, resulting in more diverse and accurate results. Recently, Pic2Word~\cite{pic2word} has been proposed to train a mapping network that converts the visual embedding into the corresponding pseudo language token, thereby improving Composed Image Retrieval (CIR) performance. Our approach is inspired by textual inversion, but we depart from not using a text encoder and instead obtain concept embeddings that are more closely aligned with the visual representation. This general technique can be applied to a variety of visual-related tasks.

\section{Method}
\label{sec:method}
Our approach can be divided into two stages: the concept embedding search stage and fine-tuning stage, where the first stage is a crucial phase that embodies our core idea. In this phase, our method focuses on searching for a set of embedding vectors representative enough to substitute the original prompts. Before we delve into the detail of our method, we need first to have a grasp of the VLMs.

\subsection{Preliminaries}
\noindent\textbf{Vision-Language Models}
The dual-stream network is a commonly adopted structure in the vision-language models, where two parallel encoders map text and image inputs into feature spaces:
\begin{equation}
\label{encoders}
V=\mathbf{E}_{\mathtt{visual}}(\text{Image}),P=\mathbf{E}_{\mathtt{text}}(\text{Prompt}).
\end{equation}
Typically, some early works, such as CLIP~\cite{clip} or ALGN~\cite{align}, align the text and image embedding spaces by giving the prediction probability as followings:
\begin{equation}
\label{eq:contrastive_loss}
    p(y=i|x) = \frac{exp(cos(v, p_i)/\tau)}{\sum_{j=1}^K{exp(cos(v, p_j)/\tau)}},
\end{equation}
where $v$ represents the feature extracted from the image encoder $\mathbf{E}_{\mathtt{visual}}$ w.r.t image $x$, $\{p_i\}_{i=1}^K$ denote a set of embeddings obtained from the text encoder for a $K$ size batch of text prompts, and $cos(\cdot, \cdot)$ is the cosine similarity, while $\tau$ is the temperature parameter. By optimizing the cosine similarity of the relevant image and text pair, VLMs can align the image and text encoder representation.

\noindent\textbf{Prompt Tuning} Different from prompt engineering, prompt tuning methods~\cite{prompt-tuning,p-tuning} prepend $m$ tunable prefix vectors to the input word embeddings in the first layer; CoOP~\cite{coop} borrowed this idea and prepend or append the tunable vectors to the class name embeddings. Concretely, the prompt embeddings combined with tunable vectors can be written in the following format:
\begin{equation}
\label{eq:p-tuning}
    P = [P]_1...[P]_{\frac{M}{2}}[Class][P]_{\frac{M}{2}+1}...[P]_M,    
\end{equation}
where $[P]_m (m \in \{1, ..., M\})$ is a tunable vector with the same dimension as the class word embedding. Then this prompt will be fed into the frozen VLM for fine-tuning these tunable vectors.

\noindent\textbf{Textual Inversion}
Textual inversion~\cite{text_inversion} is a novel technique in text-to-image generation tasks, especially in guiding personalized creation and style transfer. Textual inversion uses a new `word' token in the text embedding space to capture the specific visual concept and style from a couple of user-provided images. This `word' can be combined with context to form a text prompt, guiding the image generation process. It uses the Latent Diffusion Models (LDMs)~\cite{rombach2022high} to generate images. LDMs first use a pre-trained autoencoder to map the input image $x$ into a latent embedding $z = \sigma(x)$ and then use the diffusion method to generate images from the latent represents.
The process of textual inversion can be formulated as the following equation:
\begin{align}
\label{eq:generalization}
p_{cls}^* = \text{arg}&\min_{\bm{p_{cls}}}\mathbb{E}_{z\sim\sigma(x), t,x_0,\epsilon\sim N(0, I)}\Big[\lVert\epsilon - \epsilon_\theta(z_t,t,\mathbf{E}_{\mathtt{text}}(p_{\text{template}})\rVert^2_2 \Big]
\end{align}
where t is the time step, $z_t$ is the latent embedding at time t, $\epsilon$ denotes the random noise sampled from the Gaussian distribution and $\epsilon_\theta(\cdot)$ is the denoising network. $\mathbf{E}_{\mathtt{text}}$ is the text encoder and $p_\text{{template}} = \{P; p_{cls}\}$. $P$ is the context prompt and $p_{\text{template}}$ is the text input integrated with the extra `word' token to be trained. And $p_{cls}$ is the optimization goal, the embedding vectors for the extra `word' token which refers to the unseen class.

\begin{figure*}
  \centering
    \includegraphics[width=1.0\linewidth]{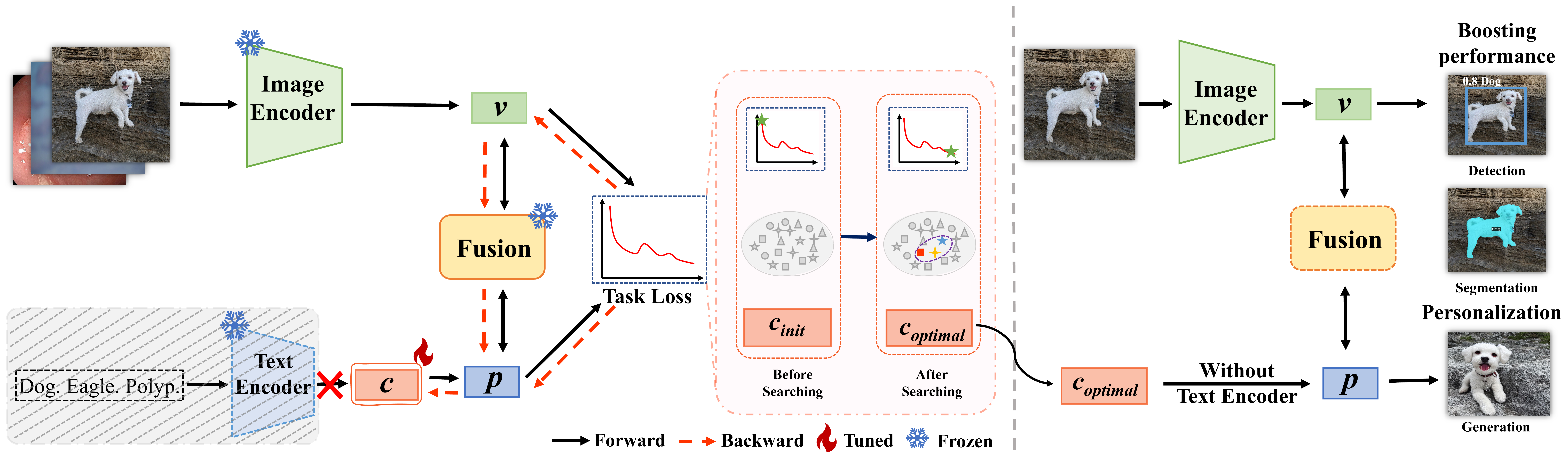}
  \caption{Method overview. Our two-stage method comprises an embedding search stage and a fine-tuning stage. In the search stage, we discard the text encoder and utilize gradient backpropagation to identify downstream task-relevant concept embeddings in the pre-trained model's latent space. In the fine-tuning stage, we leverage the concept embeddings from the first stage to improve downstream task performance, including detection and segmentation, or be utilized for personalized image generation tasks. }
  \label{fig:main}
  \vspace{-6mm}
\end{figure*}

\subsection{Concept Embedding Search}
As mentioned above, we propose a two-stage PEFT method for finding the proper concept embeddings without the text encoder, and the first stage reflects the core idea of concept embedding search. The high-level structure is illustrated in Figure \ref{fig:main}. Specifically, we initialize a fixed number of vectors to replace the text prompt inputs $P$. Formally, the concept embedding vectors can be written in the following format: $C = [C]_1[C]_2...[C]_M$, where $[C]_m (m \in \{1, ..., M\})$ is a vector with the same dimension as the text embedding vectors.
It is worth noting that the whole text encoder module will be removed from our structure, shown as the shadowed area in Figure \ref{fig:main}, and this removal will improve the tuning efficiency by a large margin. To tune the tunable vectors into concept embeddings with semantic meaning, we use appropriate task objectives to optimize these vectors while keeping the whole image encoder $\mathbf{E}_{\mathtt{visual}}$ frozen. 
As such, these searched concept embeddings will be fused with the untouched visual embeddings from the image encoder. 
In that sense, if we tune these randomly initialized vectors with target images, we will receive a set of embeddings that is deeply aligned with the visual concept of the target class. Therefore we initialize a bunch of embedding vectors for every unique class in the dataset and reserve all of these embeddings for the next stage. 

We can also train these adjustable embedding vectors with a variety of task objectives, either in an aggregated or separated manner. The above process can be formulated as the following equation:
\begin{align}
\label{eq:concept search}
    \bm{c}^{\ast} = \text{arg}&\min_{\bm{c}}\mathbb{E}_{(X, y)\sim\mathcal{D_{\text{train}}}}\sum_i^m{l_i(\text{VLM}(X, c; \theta_{\text{frozen}}), y)},
\end{align}
where $l_i \in L=\{l_{cls}, l_{bbox}, l_{mask}, l_{gen}\}, 1 \le m \le |L|$, and $\text{VLM}(\cdot)$ represents the VLM with frozen parameters $\theta_{\text{frozen}}$ that takes the image $X$ from training dataset $D_{\text{train}}$ and concept embedding $c$ as input and returns a prediction of the label $y$. The label formats can vary according to the task, and so does the loss function. 
Our target is to obtain the optimal concept embedding $c^*$. Indeed, our method can be optimized with either one single task loss, such as classification task loss, or a combination of several task losses. As we will show later, the combination strategy of the loss functions can affect the final results, but any arbitrary combination can achieve a decent result. So, we suggest selecting a proper loss function combination for different tasks and datasets.

\subsection{Embedding Empowered Performance Boosting}
Once we obtain the searched embedding vectors, we can now fix the embedding vectors’ parameters and unfreeze the rest parameters of the VLMs to fine-tune with the task loss functions:
\begin{equation}
\label{eq:stage2}
    \bm{\theta}^{\ast} = \text{arg}\min_{\bm{\theta}}\mathbb{E}_{(X, y)\sim\mathcal{D_{\text{train}}}}\sum_i{l_i(\mathcal{\text{VLM}}(X, c^{\ast};\theta), y)},
\end{equation}
where $l_i$ denotes the loss function of the downstream task and $l_i \in \{l_{cls}, l_{bbox}, l_{mask}, l_{gen} \}$ can be different losses. Note that the optimal $c^*$ obtained from the previous stage will not be tunable during this stage. In general, we use our searched embedding on different downstream tasks and show that fine-tuning with our concept embeddings can achieve better performance than the traditional fine-tuning methods. 

Moreover, our searched embedding vector can also be used for personalized image generalization for some specific classes/objects. Unlike general text-to-image generation tasks, personalized image generalization tasks require the model's deep understanding and detail-preserving ability. As we repeatedly stressed in the above passages, our method can obtain a set of embedding vectors close to the image encoder embedding space and can better capture the detail of the visual concept. Therefore, feeding our embedding to the diffusion-like generation model can inject high-quality visual concepts into the model. In this work, we follow \cite{text_inversion} using LDMs as our image generation module.

\section{Experiments}
\label{sec:exp}
In this section, we delineate the evaluation tasks and benchmarks used in this work, as well as the implementation details. Our primary results encompass three commonly used tasks: Object Detection, Instance Segmentation, and Personalized Image Generation.

\subsection{Setup}

\noindent\textbf{Pre-trained Models and Datasets.} We validate the effectiveness and applicability of our approach on three widely-used tasks. For the object detection task, we utilize the pre-trained model GLIP~\cite{glipv1} and validate our approach on a suite of 13 ODinW (object detection in the wild) datasets. In addition to natural image datasets, we also evaluate our approach on 8 datasets from the medical domain, comprising 4 non-radiology (ISIC 2016~\cite{isic2016}, DFUC 2020~\cite{dfuc2020}, BCCD, CPM-17~\cite{cpm17}) and 4 radiology (TBX11k~\cite{tbx11k}, LUNA16~\cite{luna16}, ADNI~\cite{adni}, TN3k~\cite{tn3k}) datasets. For the instance segmentation task, we use the pre-trained model UNINEXT~\cite{uninext} (from the 2nd stage) and perform validation on 1 natural image dataset (Cityscapes~\cite{cityscapes}) and 2 medical datasets (DFUC2022~\cite{dfuc2022}, Kavsir-SEG~\cite{kvasir}). For the personalized image generation task, we employ the popular and advanced pre-trained generative model, Stable Diffusion~\cite{stablediffusion} (stable-diffusion-v1-5) from Diffusers and evaluate our method on the DreamBooth~\cite{dreambooth}, which encompasses 30 subjects. For more details, please refer to the appendix.

\noindent\textbf{Implementation Details.}  For GLIP,  we use a learning rate of 1e-3, weight decay rate of 0.05, and employ 3 tokens per class for concept embedding search. We follow the transfer learning experiment settings in GLIP for other hyperparameters. During the fine-tuning stage, we set the learning rate to 1e-5. For UNINEXT, we employ a learning rate of 2e-3 and use 2 tokens per class for concept embedding search.
During the fine-tuning stage, the learning rate is set to 2e-5.  Following UNINEXT, we conduct each experiment for 12 epochs. For Cityscapes, we report the validation result from the last epoch, while for the other two datasets, we report the testing result from the best-performing model on the validation set. In the Stable Diffusion experiment, we use a learning rate of 1e-3, gradient accumulation step of 4, and train for 3000 iterations. We generate images using the final obtained embedding and 1 token per subject for the concept embedding search stage. In addition, we employ 4$\times$3090 GPUs with a total batch size of 4 to conduct the experiments of GLIP and UNINEXT, and 1$\times$3090 GPU with a batch size of 1 to perform the experiments of Stable Diffusion.

\subsection{Fine-tuning Performance on 3 Tasks}
To demonstrate the versatility of our proposed method, we conduct experiments on more than 24 datasets for 3 different tasks, including detection, segmentation and image generation. We use GLIP, UNINEXT, and LDM as our base models and follow their original settings. For each model, we only change the text prompt embeddings while keeping other components untouched. 

\noindent\textbf{Object Detection}
\input{tables/result_detection}
We compare the first stage of our method with other common PEFT methods. After the first stage finishes training the tunable vectors, we directly inject these learned vectors into the base VLM. For fair comparisons, we also fine-tune other methods with the same base model. Specifically, we use the same number of tunable embedding vectors for prompt tuning, textual inversion, and our method. Then we follow the setting in~\cite{glipv1} for linear probing, which only unfreezes the linear layer of the pre-trained model.
From Table \ref{tab:ODinW/ODinM_OD}, we can see that all four methods unfreeze about the same number of parameters, but our method has about only half of the total parameters. This lightweight structure is attributed to the removal of the text encoder and largely accelerates the process. As shown in table \ref{tab:ODinW/ODinM_OD}, our method consistently achieves superior results on more than 21 datasets compared to other tuning methods with only half of the parameters of the pre-trained model. 

For the second stage, we compare our method with the full model fine-tuning method. We feed the VLMs our tuned concept embeddings as prompt and fine-tune the whole model while keeping our embeddings unchanged. From Table~\ref{tab:ODinW/ODinM_OD}, we can see that our second stage method also receives remarkable performance on all detection datasets that we have evaluated.

Besides the superior performance, we also notice our method has a stronger generalization capability than other methods. We roughly split our collected datasets into three major groups based on the domains they belong to, including natural, medical non-radiology, and medical radiology domains. Since all of the VLMs are trained with natural images, medical data, especially radiology images, can be considered as out of the training domain for the VLMs.
As shown in the last column of Table~\ref{tab:ablation_tasks},  
the superiority of our first-stage method is extra obvious in radiology datasets, where our method has about a 5\% increase in Average Precision (AP) compared to the second-best tuning method. 
We conjecture that this outstanding performance in out of domain dataset is attributed to the close distance between our concept embeddings and the image embeddings. And due to the large domain gap between radiology images and natural images, there may exist no suitable `word' embeddings that can fully capture the visual concept of the class/object in the radiology domain. The reliance on text encoders hinders other methods from fully adapting to the unseen visual concept.

\noindent\textbf{Instance Segmentation}
We also test our method on three different datasets for instance segmentation task, using UNINEXT model as our base model for all tuning methods. The patterns in the results are similar to what we observe in the object detection task. As shown in Table \ref{tab:results1}, our method achieves comparable results in the first stage and outperforms the full model fine-tuning on all datasets. Moreover, we also discover our method's generalization capability. As presented in Table \ref{tab:results1}, our method’s performance surpasses other methods by a larger margin on the DFUC2022 and Kavisr-SEG datasets, which contained images in the medical domain.

\input{tables/result_seg}
\begin{figure*}
  \centering
    \includegraphics[width=1.0\linewidth]{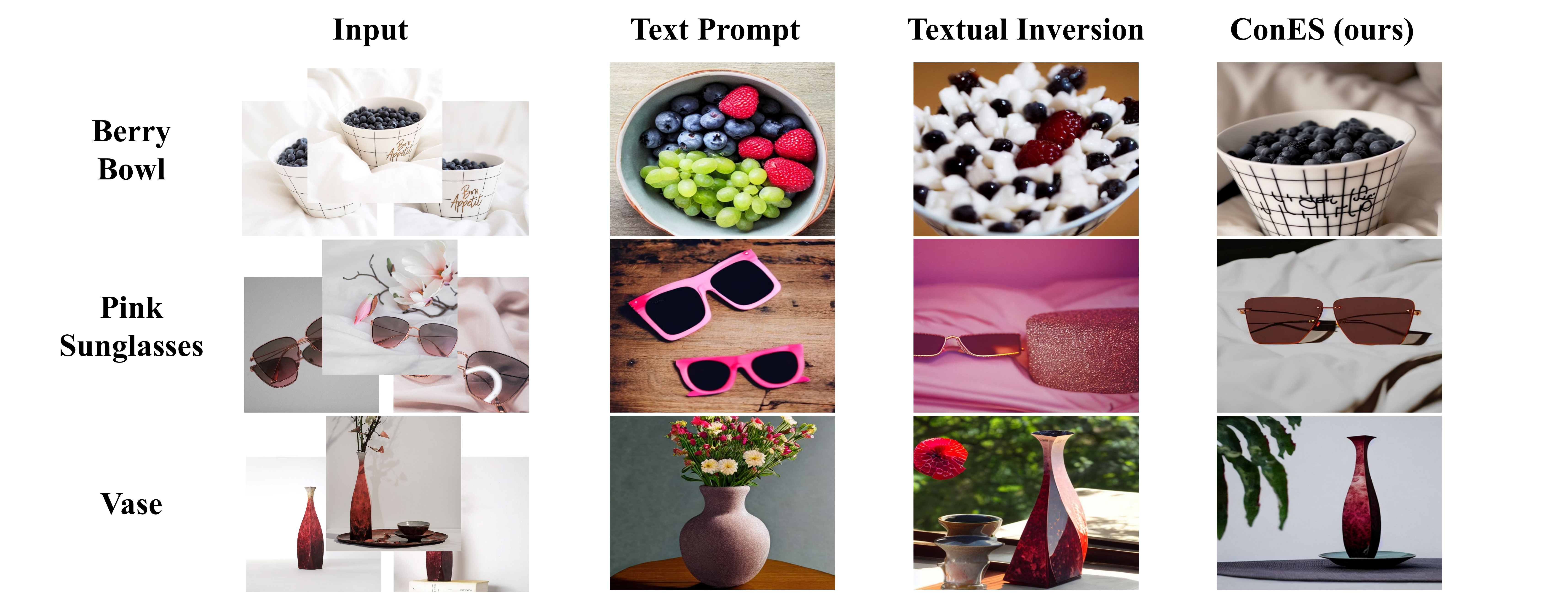}
  \caption{Illustrative examples of personalized image generation results are shown. For image generation in the original Stable diffusion, the `A photo of [CLS]' text prompt is used, while `A photo of *' is employed for textual inversion and our proposed ConES.}
  \label{fig:generation}
  \vspace{-6mm}
\end{figure*}

\noindent\textbf{Personalized Image Generation}
As mentioned before, we believe our learned vectors are ideal prompts for guiding the image generation model to generate objects with specific visual features. This work uses the LDM model to generate images from our prompts. We show several qualitative examples of our generated images to show our embeddings can capture more visual details than other methods. In Figure \ref{fig:generation}, we compare the ability to capture and create variations of an object with specific details from 4-6 image inputs. Concretely, we try three methods to capture the input images into prompts: (1) The first method is straightforward. We simply use the object name words as the text prompt to feed into the text encoder of the CLIP model and use such text embedding to guide the image generation model; (2) For the second method, we simply follow the textual inversion algorithm in ~\cite{text_inversion} to represent the provided visual concept into a single `word'. Then we combine such `word' with text templates to obtain the text embeddings; (3) We use the first stage method of our framework to train a fixed number of embedding vectors and directly use such vectors to guide the LDM, without going through the text encoder.
As illustrated in Figure \ref{fig:generation}, variants generated by our methods restore the visual concept more accurately. And images generated by other methods are less controlled by the original style, because their prompts can not fully capture the visual concept.

\subsection{Visualization of the Concept Embedding Spaces}
\begin{figure*}
  \centering
   \vspace{-6mm}
    \includegraphics[width=0.92\linewidth]{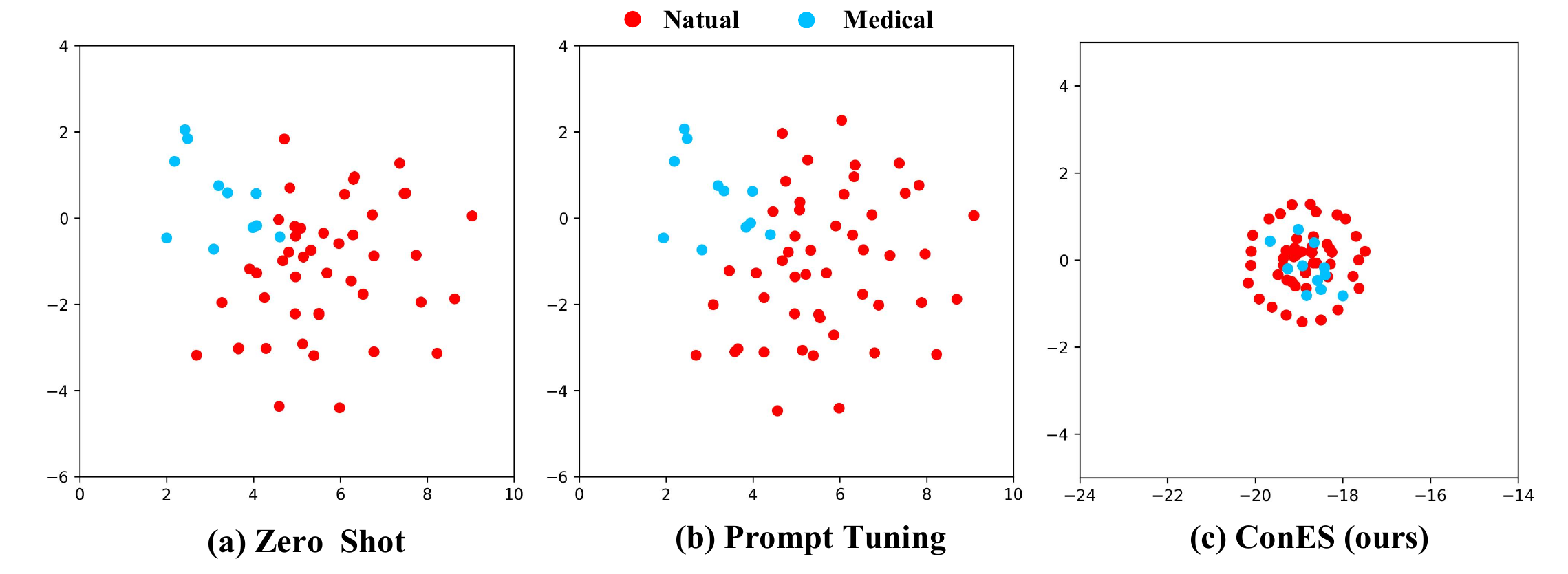}
    \vspace{-2mm}
  \caption{t-SNE visualizations of the text embeddings $\mathbf{p}$ obtained with Text Encoder and the concept embeddings $\mathbf{c}$ obtained without Text Encoder through our proposed concept embedding search.} 
  \vspace{-4mm}
  \label{fig:tsne_text}
\end{figure*}

\noindent\textbf{Concept Embedding Distribution} In Figure \ref{fig:tsne_text}, we visualize the distribution of concept embeddings of every class for all natural and medical image datasets. As illustrated in the figure, the red dots represent the natural class embeddings, and the blue dots represent the medical class embeddings. The distribution in Figure \ref{fig:tsne_text} (a) and (b) are generated by the text encoder of previous methods, and we observe that they follow the same pattern. Both of the figures show that the medical and the natural concepts are segregated into two separate clusters. This pattern suggests that  unseen medical concepts can not be interpreted with previously acquired concepts. Therefore these unseen concepts stay in the edge area of the text encoder embedding space and can not be integrated into the distribution even after prompt-tuning, as shown in Figure \ref{fig:tsne_text} (b). Oppositely, Figure \ref{fig:tsne_text} (c) shows the concept embedding distribution generated by our method.
As one can tell, the medical concepts and natural concepts are well-blended into the same cluster after training. 
Though medical and natural concepts are in different contexts in language, their visual elements could be from the same distribution. 
Since the image encoder has seen these visual elements in some natural concepts before, so the unseen concepts won’t shift apart from the visual embedding space. A more straightforward analogy is that a vision model might recognize a cell as donuts, but a language model can only relate unseen concepts to morphologically related words. The above discussion partially explains the stronger generalization capability of our methods on unseen concepts.

\begin{figure*}
  \centering
    \includegraphics[width=0.92\linewidth]{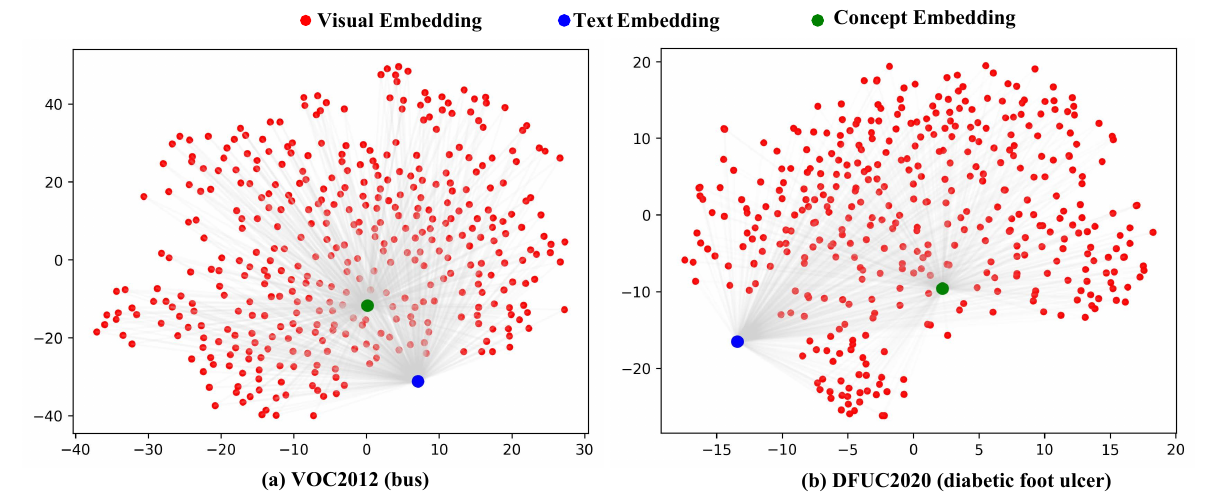}
    \vspace{-2mm}
  \caption{t-SNE visualizations of the 400 Visual Embeddings (red), Text Embedding (blue), and Concept Embedding (green), all generated by the projection value linear layer before the first early fusion of the GLIP-T.} 
  \label{fig:tsne_vt}
  \vspace{-6mm}
\end{figure*}

\noindent\textbf{Modality Gap in the Embedding Space}
Figure \ref{fig:tsne_vt} further confirms our conjecture by putting embeddings obtained by our method, visual embeddings from the image encoder, and text embeddings from the text embedding altogether. Concretely, the red dots in the figures represent the visual embeddings obtained from the image encoder for images in the same class. The green dot is the concept embedding of such a class obtained by our method, and the blue dot is the text embedding from the text encoder by providing the class name.
Figure \ref{fig:tsne_vt} (a) and (b) demonstrate the embedding space of a class from a natural image dataset and a class from a medical domain dataset, respectively. It is obvious that the embedding tuned by our method is closer to all of the image embeddings on average than the text embedding, as shown in Table \ref{tab:distances}. Thus, we can conclude that our method indeed mitigates the modality gap to learn a comprehensive representation of the visual concepts.

\input{tables/distance}

\input{tables/ablation_tasks}

\subsection{Ablation Studies and Discussions}
We conduct two ablation studies to study the effect of the number of embedding vectors and the selection of loss function.

\noindent\textbf{Loss Function Selection} For the first stage of our framework, we can use different combinations of loss functions to tune the random initialized vectors. For example, we show the relationship between the choice of loss functions and the instance segmentation results obtained by applying the embedding vectors tuned with the chosen loss functions on the Kavisr-SEG dataset in Table \ref{tab:ablation_tasks}. We argue that adding classification loss into the loss function is crucial for training a successful concept embedding, and our method can be trained with different loss functions to receive a decent result.

\noindent\textbf{Number of Embedding Vectors} We also study the effect of using different numbers of tunable vectors. As shown in Table \ref{tab:ablation_tokens}, multiple tokens are necessary to obtain effective embeddings of the concepts. For concept within the same domain of pre-training, the required number of tokens may be less than the number of tokens required for out of domain concepts. In Table \ref{tab:ablation_tokens}, the results on the Aquarium dataset, a natural image dataset, come to the peak when using three tokens. However, the results on the DFUC2020 dataset, a medical domain dataset, keep increasing when using five tokens. This pattern suggests that out-of-domain concepts or concepts for specialized domains may be more abstract compare to the in-domain concepts and thus require more tokens to capture their meaning.

\section{Conclusion}
In this work, we propose a two-stage concept embedding search approach without the use of the text encoder in the pre-trained VLMs and prove that our searched embeddings can capture the visual concept more efficiently than other prompt-tuning methods. In the first stage, we obtain the concept embedding vectors by tuning with different loss functions while keeping other components of the VLMs frozen. In the second stage, we fix the searched optimal embedding to further fine-tune the VLMs to boost the final results for different tasks, including object detection, instance segmentation, and personalized image generation. Extensive experiments show that our method can effectively learn the concept embedding and outperform other tuning methods on various tasks and datasets. From the visualization of the embedding spaces, we prove our conjecture that the embeddings learned with our method are closer to the image embeddings. Besides the advantages mentioned above, our method also halves the scale of parameters by discarding the text encoder. This lightweight structure makes our method inference faster than other PEFT methods. 
We hope our approach can provide the deep learning community with an alternate tuning paradigm for large pre-trained VLM models and expect our tuning method can be applied to more scenarios and tasks.

\newpage

\bibliographystyle{unsrt}
\bibliography{neurips_2023}

\newpage
\input{neurips_appendix}
\end{document}

%% file: tables/result_detection.tex
\begin{table}[t]
\vspace{-8mm}
\caption{Results on object detection datasets. Results with more details are provided in the Appendix.
\label{tab:ODinW/ODinM_OD}}
\setlength{\tabcolsep}{3pt}
\begin{center}
\small
\def \arraystretch{0.8}
\begin{tabular}{ccclll}
 \toprule
\multirow{2}{*}{\textbf{Method}}  & \multirow{2}{*}{\textbf{Model}} &  \multirow{2}{*}{\textbf{\tabincell{c}{\#Parameters \\ Unfrozen/All}}} &  \multicolumn{1}{c}{\textbf{Natural}} & \multicolumn{2}{c}{\textbf{Medical}} \\

\cmidrule(lr){4-4} \cmidrule(lr){5-6} 
    &    &  &  \scriptsize \textbf{ODinW(13)}  & \scriptsize \textbf{Non-Radiology(4)} & \scriptsize \textbf{Radiology(4)}  \\
 \midrule
 \multicolumn{6}{l}{\scriptsize  \demph{ \it{Zero shot} } } \\
 -  & GLIP-T & -                         & 45.6 & 6.5 & 1.4  \\
 -  & GLIP-L & -                         & 52.1 & 9.7 & 2.1 \\
\midrule
\multicolumn{6}{l}{\scriptsize \demph{ \it{Parameter efficient fine-tuning}} }\\
 Linear Probing     & GLIP-T & 0.20/231.76M     & 55.1 & 36.8  & 12.0  \\
 Textual Inversion   &  GLIP-T & 0.20/232.96M    & 46.3 & 42.1 & 24.4 \\
 Prompt Tuning     & GLIP-T & 0.20/231.96M     & 62.4 & 52.3 & 36.3 \\
 
 ConES (ours)  & GLIP-T & 0.20/123.06M     & \textbf{63.8(\textit{$\uparrow_{1.4}$})} & \textbf{52.9}(\textit{$\uparrow_{0.6}$})  & \textbf{41.3}(\textit{$\uparrow_{5.0}$}) \\ 

\cdashline{1-6}
 Linear Probing      & GLIP-L & 0.20/430.40M      & 59.2 & 38.3  & 12.8 \\
 Textual Inversion   & GLIP-L & 0.20/430.60M      & 59.2 & 45.6 & 25.1 \\
 Prompt Tuning     & GLIP-L & 0.20/430.60M      & 67.9 & 52.3 & 32.3 \\
 ConES (ours)  & GLIP-L & 0.20/321.71M     & \textbf{68.2(\textit{$\uparrow_{0.3}$})} & \textbf{52.6}(\textit{$\uparrow_{0.3}$}) & \textbf{37.5}(\textit{$\uparrow_{5.2}$}) \\ 

\midrule
\multicolumn{6}{l}{ \scriptsize \demph{ \it{Fine-tuning} }} \\
 Full Model            & GLIP-T & 231.15/231.76M   & 64.9 & 54.2 &  47.4\\
 ConES (ours)      & GLIP-T & 122.56/123.06M     & \textbf{65.3 (\textit{$\uparrow_{0.4}$})}& \textbf{55.5}(\textit{$\uparrow_{1.3}$}) & \textbf{48.3}(\textit{$\uparrow_{0.9}$}) \\ 

\cdashline{1-6}
 Full Model             & GLIP-L & 429.20/430.40M  & 68.9 & 56.6  &  47.3\\
 ConES (ours)      & GLIP-L & 320.03/321.71M     & \textbf{69.7(\textit{$\uparrow_{0.8}$})} & \textbf{57.1}(\textit{$\uparrow_{0.5}$}) & \textbf{49.2}(\textit{$\uparrow_{1.9}$})\\ 
\midrule

\end{tabular}

\end{center}
\vspace{-8mm}
\end{table}

%% file: tables/result_seg.tex
\begin{table}[t]
\caption{Results on instance segmentation datasets. Here, the visual backbone used in UNINEXT is ResNet50.
\label{tab:results1}}
\setlength{\tabcolsep}{3pt}
\begin{center}
\small
\def \arraystretch{0.98}
 \begin{tabular}{ccccccccccccc} 
 \toprule
    \multirow{2}{*}{\bf Method}  & \multirow{2}{*}{\bf Model} &  \multirow{2}{*}{\textbf{\tabincell{c}{\#Parameters \\ Unfrozen/All}}}  &  \multicolumn{2}{c}{\bf Cityscapes} & \multicolumn{2}{c}{\bf DFUC2022} &  \multicolumn{2}{c}{\bf Kavisr-SEG}  \\ 
    \cmidrule(lr){4-5} \cmidrule(lr){6-7} \cmidrule(l){8-9}
    
     & & &\scriptsize \bf AP$^{\text{box}}$ &\scriptsize \bf AP$^{\text{mask}}$ &\scriptsize \bf AP$^{\text{box}}$ &\scriptsize \bf AP$^{\text{mask}}$  &\scriptsize \bf AP$^{\text{box}}$ &\scriptsize \bf AP$^{\text{mask}}$ \\ 
     
     \midrule
     
       \multicolumn{9}{l}{\scriptsize{  \demph{ \it{Zero-shot} }} }\\
       
    -                    & UNINEXT & -  & 17.9    & 14.0 & 0.4 & 0.4 & 0.3 & 0.2 \\
    
    \midrule

    \multicolumn{9}{l}{\scriptsize { \demph{ \it{Parameter efficient fine-tuning} }} }\\
    
    Textual Inversion             & UNINEXT  & 0.20/166.36M   & 32.8 & 28.6 & 33.6 & 29.1  & 49.6 & 49.9 \\
    Prompt Tuning              & UNINEXT  & 0.20/166.56M   & \textbf{34.2} & 30.2 & 40.8 & 36.1  & 63.0 & 61.8 \\
    ConES (ours)           & UNINEXT  & 0.20/57.67M    & 34.1 & \textbf{30.3} & \textbf{42.1} & \textbf{37.3}  & \textbf{64.3} & \textbf{63.2} \\

    \multicolumn{9}{l}{\scriptsize { \demph{ \it{Fine-tuning} }} }\\
    Full Model         & UNINEXT  & 166.14/166.36M  & 41.3 & 33.9 & 55.3 & 52.7  & 66.6 & 69.9 \\
    ConES (ours)      & UNINEXT  & 57.25/57.67M   & \textbf{41.7} & \textbf{34.5} & \textbf{57.0} & \textbf{53.3}  & \textbf{73.5} & \textbf{74.2} \\
  \midrule

\end{tabular}

\end{center}
\vspace{-6mm}
\end{table}

%% file: tables/distance.tex
\begin{table}[t]
\caption{The average distance between Text/Concept and Visual embedding in tSNE space on 5 datasets.
\label{tab:distances}}
\setlength{\tabcolsep}{3pt}
\begin{center}
\small
\def \arraystretch{0.98}
 \begin{tabular}{ccccccc} 
 \toprule
    {\bf Method}  & {\bf VOC2012}  & {\bf DFUC2020} & {\bf CPM-17} & {\bf TBX11k}
     & {\bf TN3k} & {\bf Avg.}\\
     \midrule
     
    Zero shot      & 77.47  & 65.62   & 263.09 & 58.02 & 69.65 & 106.77 \\ 
  
    Prompt tuning      & 77.47  & 66.98   & 265.43 & 60.23 & 67.51 & 107.524 \\ 
 
     Textual inversion      & 77.58  & 67.69   & 266.6 & 61.35 & \textbf{66.48} & 107.94 \\ 

    ConES (ours)      & \textbf{77.01}  & \textbf{53.43}  & \textbf{190.74} & \textbf{37.94} & 68.12 & \textbf{85.448} \\ 
  \midrule

\end{tabular}

\end{center}
\vspace{-4mm}
\end{table}

%% file: tables/ablation_tasks.tex
\begin{minipage}{\textwidth}
\centering
\begin{minipage}[t]{0.48\textwidth}
\makeatletter\def\@captype{table}
\caption{Ablation study for task loss(Kavisr-SEG).}
\renewcommand\arraystretch{1.085}
\resizebox{\linewidth}{!}{
 \begin{tabular}{cccccc} 
 \toprule
   \bf Model  & \bf L$_{\text{CLS}}$ & \bf L$_{\text{BOX}}$ & \bf L$_{\text{MASK}}$  & \bf AP$^{\text{box}}$ &  \bf AP$^{\text{mask}}$ \\
 \midrule
  \multirow{7}{*}{\textbf{UNINEXT}} & \checkmark  &    &      & 61.6    & 61.6 \\ 
  &  & \checkmark &    &    60.1    & 58.7  \\
  &  &  & \checkmark &    49.1   & 49.3 \\
  &\checkmark  & \checkmark &     & 63.2    & 61.6 \\
  &\checkmark  &  & \checkmark     & 61.9    & 62.2  \\
  &  &\checkmark  & \checkmark    & 59.6    & 59.1  \\

  &\checkmark  &\checkmark  &\checkmark     & \textbf{64.3}    & \textbf{63.2}  \\

 \midrule

\end{tabular}}
\label{tab:ablation_tasks}
\end{minipage}
\begin{minipage}[t]{0.48\textwidth}

\makeatletter\def\@captype{table}
\caption{Ablation study of number of tokens.}
\resizebox{\linewidth}{!}{
 \begin{tabular}{cccccc} 
 \toprule
  \multirow{2}{*}{\textbf{Model}} & \multirow{2}{*}{\textbf{Tokens}} &  \multicolumn{2}{c}{\bf Aquarium}  &  \multicolumn{2}{c}{\bf DFUC2020} \\

     \cmidrule(lr){3-4} \cmidrule(lr){5-6} 

     & &  \bf AP &  \bf AP$^{\text{50}}$ &  \bf AP &  \bf AP$^{\text{50}}$ \\
 \midrule      
  \multirow{5}{*}{\textbf{GLIP-T}} & 1 & 41.1  & 67.9    & 47.1    & 80.2 \\ 
 & 2 & 51.7  & 81.0    & \textbf{49.7}    & 82.8  \\
 & 3 & \textbf{53.4}  & \textbf{83.6}    & \textbf{49.7}    & 82.3 \\ 
 & 4 & 51.4  & 82.4    & 49.5    & 83.1  \\  
 & 5 & 51.8  & 83.4    & \textbf{49.7}    & \textbf{84.1} \\ 

 \midrule

\end{tabular}}

\label{tab:ablation_tokens}
\end{minipage}
\end{minipage}

%% file: neurips_appendix.tex
\appendix
\section{Appendix}
\subsection{Model and Dataset Details.}
In this section, we present the composition detailed of every pre-trained model and dataset we collected. 

The study uses different pre-trained models~\cite{glipv1,uninext,stablediffusion} for evaluation of detection, instance segmentation, and personalized image generation tasks, and their corresponding GitHub repositories are listed in Table~\ref{table:ap_model}.

This study collects 25 open-source datasets from the internet and provides their details in Table~\ref{table:ap_dataset}. The Detection datasets are categorized into natural and medical datasets, with medical datasets further classified into non-radiology and radiology datasets. The zero-shot results in Table~\ref{tab:ODinW/ODinM_OD} demonstrate a growing domain gap from natural to non-radiology to radiology. For natural datasets, the study uses 13 ODinW datasets collected in real-world scenes, adopting dataset division settings from GLIP~\cite{glipv1}. The non-radiology datasets comprise ISIC 2016~\cite{isic2016}, DFUC 2020~\cite{dfuc2020}, BCCD, and CPM-17~\cite{cpm17}, while the radiology datasets include TBX11k~\cite{tbx11k}, LUNA16~\cite{luna16}, ADNI~\cite{adni}, and TN3k~\cite{tn3k}. The segmentation datasets are divided into natural (Cityscapes~\cite{cityscapes}) and medical (Kavisr-SEG~\cite{kvasir}, DFUC2022~\cite{dfuc2022}) datasets. The study uses DreamBooth~\cite{dreambooth} dataset for personalized imagegeneration, consisting of 30 subjects such as backpacks, stuffed animals, dogs, cats, sunglasses, cartoons, etc., with 21 objects and 9 live subjects/pets. The Cityscapes dataset is included as a reference dataset, with established public baselines, and the study reports on the validation set for this dataset. For CPM-17, the training dataset is split into training and validation datasets (8:2) since no official validation sets are available.

\subsection{Ablation Study on Deep Fusion and Pre-training data}\label{sec:ablation}
There are two types of fusion strategies for VLMs: 1. deep(early) fusion or 2. late fusion. As discussed above, the embedding vectors from the text encoder $E_{\text{text}}$ and the visual encoder $E_{\text{image}}$ were parallel initially, and their fusion results, such as Equation~\ref{eq:contrastive_loss}, will give us cross-modal representation. However, in some VL literature \cite{glipv1, glipv2}, this fusion with only a dot product at the last stage is called late fusion, and they introduce deep(early) fusion via a cross-modality fusion module based on a cross-attention mechanism between encoders at an earlier phase.  

Concretely,  a deep fusion operation will be applied to the last few layers of both encoders by using multi-head cross attention between two encoders. The fusion process can be formulated as follows:
after obtaining the independent image and text embeddings,
\begin{align}
\label{eq:fusion}
    V_{i2t}, P_{t2i} = \text{XAttn}(V, P); 
    V' = V + V_{i2t}; P' = P + P_{t2i},
\end{align}
where $\text{XAttn}(\cdot, \cdot)$ denotes the general cross-modality attention module. $V_{i2t}, P_{t2i}$ represent the image-to-text and text-to-image context vectors obtained from the cross-modality module. The cross-modality fusion is completed by adding the context vectors to the original feature embeddings. We also observed that the concept embedding vectors searched by our proposed approach work better with the VLMs with the early-fusion operation. As shown in Table \ref{tab:earlyfusion}, we can see the inferior performance of the searched embedding in the first row, where we removed the cross-attention module from the GLIP model.

Besides the deep fusion factor, Table \ref{tab:earlyfusion} also shows a trend of adding pre-train data size will increase the model's generalization capability. Specifically, the GLIP model trained with O365 data set show inferior performance on DFUC2020 and TBX11k dataset, because those datasets contains unseen concepts. However, the the GLIP variant trained with extra data perform better on those out of domain datasets, compared to the GLIP variant trained with only O365 dataset.

\subsection{More Detailed Display of Object Detection Results}
In this section, we show the detail results for each dataset in the ODinW and medial datasets. We only show the average results in the experiments section. As shown in Table \ref{table:zero_shot_full}, we show the detail results for two variants of GLIP model using different tuning methods for 13 datasets. In Table \ref{table:zero_shot_full_med}, we show the detail results for 8 medical datasets, including various medical modality. 

\subsection{Supplementary Personalized Image Generation Results }\label{sec:additional_results}
\begin{figure}[b]
  \centering
    \includegraphics[width=1.0\linewidth]{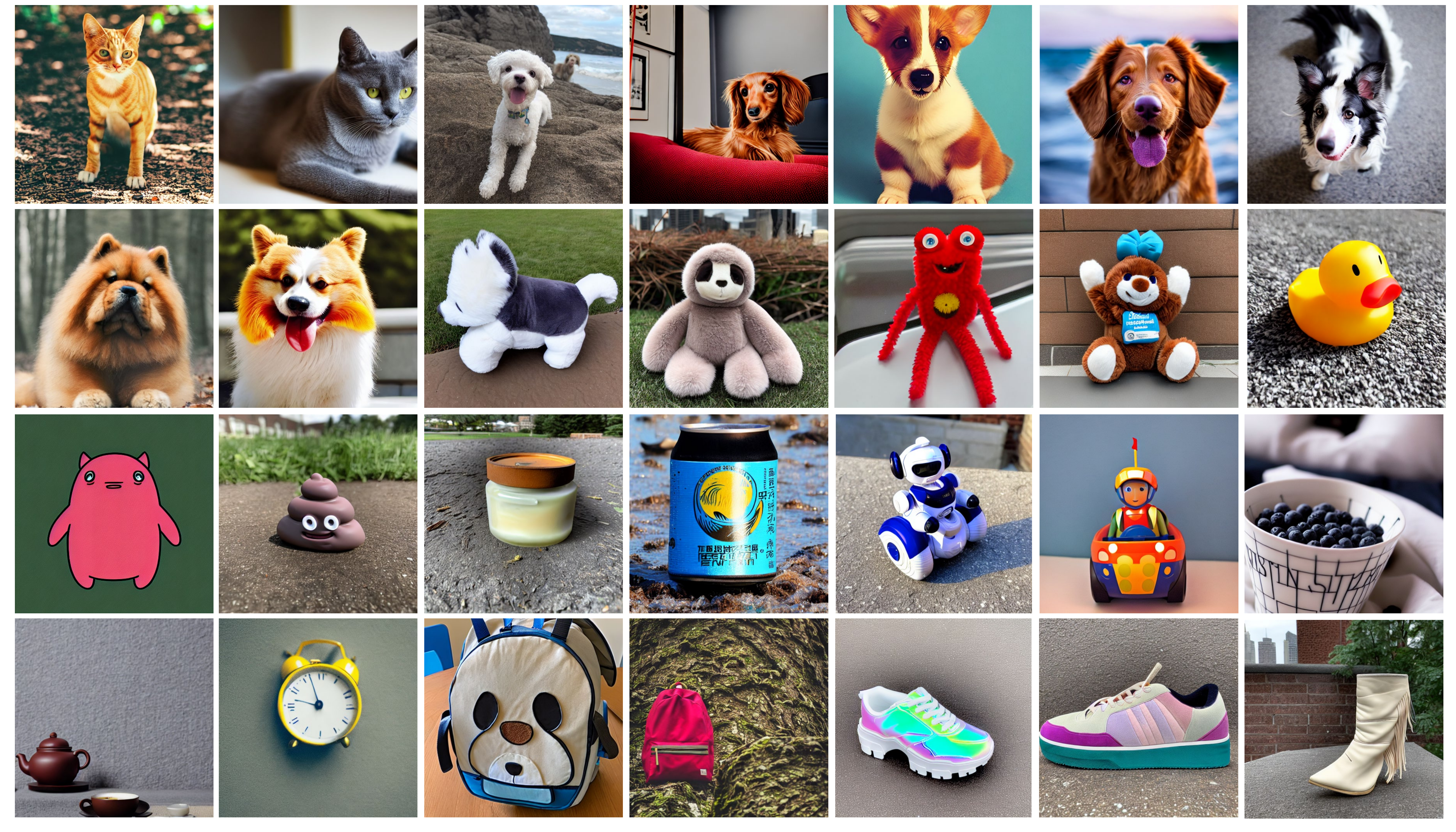}
  \caption{Additional illustrative examples of personalized image generation results trained on the DreamBooth dataset are shown.}
  \label{fig:generation}
  \vspace{-2mm}
\end{figure}

In this section, we will exhibit some extra qualitative examples of the generated image using our searched embeddings. As illustrated in Figure~\ref{fig:generation}, we show various images learned from different concepts. Though there still remains room for improvement, our generated images are comparable to some SOTA models already.

\input{tables/appendix_models}

\input{tables/appendix_datasets}

\input{tables/odinw}

\input{tables/appendix_ablation_early_fusion}


\input{tables/medical}

%% file: tables/appendix_models.tex
\begin{table*}[ht]
\caption{Description of the pre-trained model utilized in the study. \label{table:ap_model}}
\setlength{\tabcolsep}{2pt}
\begin{center}

\resizebox{\linewidth}{!}{
\begin{tabular}{l@{\hskip9pt} | 
c@{\hskip9pt}|c@{\hskip9pt}c@{\hskip9pt}c@{\hskip9pt}c@{\hskip9pt}c@{\hskip9pt}c@{\hskip9pt}c@{\hskip9pt}c@{\hskip9pt}}
\toprule

 Model &  Task  &  URL \\
\midrule
 GLIP~\cite{glipv1} &  Detection  &  \url{https://github.com/microsoft/GLIP} \\
 UNINEXT~\cite{uninext}  &  Instance Segmentation  &   \url{https://github.com/MasterBin-IIAU/UNINEXT}\\
 Stable  Diffusion~\cite{stablediffusion} &  Generation &  \url{https://github.com/huggingface/diffusers} \\

\bottomrule
\end{tabular}
}
\end{center}
\end{table*}

%% file: tables/appendix_datasets.tex
\begin{table*}[ht]
\caption{Description of the datasets utilized in the study. \label{table:ap_dataset}}
\setlength{\tabcolsep}{3pt}
\begin{center}
\resizebox{\linewidth}{!}{
\begin{tabular}{l@{\hskip9pt} | 
c@{\hskip9pt}|c@{\hskip9pt}|
c@{\hskip9pt} c@{\hskip9pt}c@{\hskip9pt}
c@{\hskip9pt}c@{\hskip9pt}c@{\hskip9pt}  c@{\hskip9pt}
c@{\hskip9pt}c@{\hskip9pt}c@{\hskip9pt}c@{\hskip9pt}c@{\hskip9pt}c}
\toprule

Dataset & Class & Train/Val/Test & URL \\
\midrule
ODinW(13) & 50 (total) & -/-/- & \url{https://github.com/microsoft/GLIP} \\
ISIC2016~\cite{isic2016}  & 2 & 720/180/379 &  \url{https://challenge.isic-archive.com/data}\\
DFUC2020~\cite{dfuc2020} & 1 & 448/127/63 & \url{https://dfu-challenge.github.io} \\
BCCD  & 3 & 765/73/36 & \url{https://public.roboflow.com/object-detection/bccd} \\
CPM-17~\cite{cpm17} & 1 & 25/7/32 & \url{https://github.com/vqdang/hover_net} \\
TBX11k~\cite{tbx11k} & 1 & 479/120/200 &  \url{https://mmcheng.net/tb}\\
Luna16~\cite{luna16} & 1 & 759/190/237 & \url{https://luna16.grand-challenge.org/Data} \\

ADNI~\cite{adni} & 1 & 2590/589/818 & \url{https://www.kaggle.com/datasets/sabermalek/mrihs} \\

TN3k~\cite{tn3k} & 1 & 2303/576/614 & \url{https://github.com/haifangong/TRFE-Net-for-thyroid-nodule-segmentation} \\

Cityscapes~\cite{cityscapes} & 20 & 2975/500/- & \url{https://public.roboflow.com/object-detection/vehicles-openimages} \\

DFUC2022~\cite{dfuc2022} & 1 & 1280/320/400 & \url{https://dfu-challenge.github.io}\\

Kavisr-SEG~\cite{kvasir} & 1 & 718/182/100 &  \url{https://datasets.simula.no/kvasir-seg} \\

DreamBooth~\cite{dreambooth} & 30 & 4 - 6 per class & \url{https://github.com/google/dreambooth} \\

\bottomrule
\end{tabular}
}
\end{center}
\end{table*}

%% file: tables/odinw.tex
\begin{table*}[ht]
\caption{Object detection results on ODinW(13) datasets(AP\%).}
\label{table:zero_shot_full}
\begin{center}
\resizebox{\linewidth}{!}{
\begin{tabular}{l@{\hskip9pt}l@{\hskip9pt}| 
c@{\hskip9pt}c@{\hskip9pt}c@{\hskip9pt} 
c@{\hskip9pt}c@{\hskip9pt}c@{\hskip9pt}
c@{\hskip9pt}c@{\hskip9pt}c@{\hskip9pt}c@{\hskip9pt}
c@{\hskip9pt}c@{\hskip9pt}c@{\hskip9pt}c@{\hskip9pt}c@{\hskip9pt}c@{\hskip9pt}c}
\toprule

Model  & Method & \small{PascalVOC} &
\small{AerialDrone} & 
\small{Aquarium} &
\small{Rabbits} &
\small{EgoHands} &
\small{Mushrooms} &
\small{Packages} &
\small{Raccoon} &
\small{Shellfish} &
\small{Vehicles} &
\small{Pistols} &
\small{Pothole} &
\small{Thermal} & 
Avg \\ \midrule
\multirow{7}{*}{GLIP-T} & Zero-shot               & 56.2          & 12.5          & 18.4          & 70.2          & 50.0          & 73.8          & 72.3          & 57.8          & 26.3          &56.0          & 49.6          & 17.7          & 44.1          & 46.5 \\
                        & Linear Probing          & 65.5          & 14.1          & 36.5          & 68.2          & 67.2          & 76.6          & 70.2          & 63.8          & 29.1          & 65.5          & 63.5          & 29.9          & 66.5          & 55.1 \\
                        & Textual Inversion & 66.7          & 8.0          & 4.5          & 69.3          & 71.1          & 62.9          & 18.4          & 62.1          & 31.2          & 57.8          & 68.0          & 40.1          & 41.4          & 46.3 \\
                        & Prompt Tuning          & 66.4          &\textbf{ 27.6}          & 50.9          &\textbf{ 70.6}          & 73.3          & \textbf{88.1} & 67.7          & \textbf{64.0}          & 40.3          &\textbf{ 65.4}          & 68.3          & 50.7          & 78.5          & 62.4 \\
                        & ConES (ours)               &\textbf{ 69.3 }         & 22.0          &\textbf{ 53.4}          & 70.1          & \textbf{75.7 }         & 87.5          &\textbf{ 73.9}          & \textbf{64.0}          & \textbf{44.3}          & 64.7          &\textbf{ 71.8}          & \textbf{52.4}          & \textbf{80.3} & \textbf{63.8 }\\
\cdashline{2-16}
                        & Full Model            & 62.3          & \textbf{31.2} & 52.5          & \textbf{70.8} & \textbf{78.7} & \textbf{88.1} & 75.6          & 61.4          & \textbf{51.4} & 65.3          & 71.2          & \textbf{58.7} & 76.7          & 64.9 \\

                        & ConES (ours)         & \textbf{69.8} & 24.6          & \textbf{54.3} & 70.0          & 77.5          & \textbf{88.1} & \textbf{76.2} & \textbf{66.4} & 45.2          & \textbf{66.1} & \textbf{72.9} & 58.0          & \textbf{80.0}          &\textbf{ 65.3 }\\ \midrule
\multirow{7}{*}{GLIP-L} & Zero-shot         & 61.7          & 7.1          & 26.9          & 75.0          & 45.5          & 49.0          & 62.8          & 63.3          & 68.9          & 57.3          & 68.6          & 25.7          & 66.0          & 52.1 \\
                        &  Linear Probing            & 70.9          & 9.6          & 42.3          & \textbf{75.3 }         & 70.5          & 39.4          & 69.3          & \textbf{71.6} & \textbf{73.9} & 69.7          & 72.1          & 33.2          & 72.3          & 59.2 \\
                        & Textual Inversion & \textbf{75.0} &9.1          & 51.8          & 71.3          & 74.5          & 42.4          & 67.0          & 68.7          & 68.9          & 71.0          & 73.7          & 45.0          & 51.5          & 59.2 \\
                        & Prompt Tuning          & 72.9          & 23.0          & 51.8          &72.0          & 75.8          & \textbf{88.1} & \textbf{75.2}          & 69.5          & 73.6          & \textbf{72.1}          & 73.7          & 53.5          & 81.4          & 67.9 \\
                        & ConES (ours)               & \textbf{75.0} & \textbf{26.6 }         & \textbf{55.2}          & 71.9          &\textbf{ 77.2 }         & 87.5          & 74.3          & 66.4          & 70.5          & 70.3          & \textbf{74.0}          &\textbf{ 54.6  }        &\textbf{ 83.0}          & \textbf{68.2} \\
\cdashline{2-16}
                        & Full Model      & 69.6          & \textbf{32.6} & 56.6          & \textbf{76.4} & \textbf{79.4} & \textbf{88.1} & 67.1          & 69.4          & 65.8          & 71.6          & \textbf{75.7} & \textbf{60.3} & 83.1          & 68.9 \\

                        & ConES (ours)              & \textbf{75.0} & 27.9          & \textbf{58.3} & 72.0          & 78.7          & \textbf{88.1} & \textbf{77.2} & 69.9          & 68.1          & \textbf{72.4} & 75.0          & 58.9          & \textbf{84.6} & \textbf{69.7} \\

\bottomrule
\end{tabular}
}
\end{center}
\end{table*}

%% file: tables/appendix_ablation_early_fusion.tex
\begin{table}[ht]
\caption{Ablation study on the pre-training data and the impact of deep fusion strategy(AP\%).
\label{tab:earlyfusion}}
\vspace{2mm}
\setlength{\tabcolsep}{3pt}
\begin{center}
\small
\def \arraystretch{0.98}
 \begin{tabular}{cccccc} 
 \toprule
\bf Model  & \bf Deep Fusion &\bf Pre-train Data  &\bf Aquarium  &\bf DFUC2020  &\bf TBX11k\\
 
     \midrule

       \multirow{4}{*}{\textbf{GLIP-T}}         & \xmark  & O365  & 31.5 & 12.5 & 2.6  \\
        & \cmark  & O365    &50.0 & 45.9 & 30.4  \\
        & \cmark  & O365, GoldG    &52.0  & \textbf{49.8} & 32.6 \\
        & \cmark  & O365, GoldG, CC3M, SBU    &\textbf{53.4} & 49.7 & \textbf{33.4}  \\

  \midrule

\end{tabular}

\end{center}

\end{table}

%% file: tables/medical.tex
\begin{table*}[ht]
\caption{Object detection results on 8 medical datasets(AP\%).}
\label{table:zero_shot_full_med}
\begin{center}
\resizebox{\linewidth}{!}{
\begin{tabular}{l@{\hskip9pt}l@{\hskip9pt}| 
c@{\hskip9pt}c@{\hskip9pt}c@{\hskip9pt} 
c@{\hskip9pt}c@{\hskip9pt}c@{\hskip9pt}
c@{\hskip9pt}c@{\hskip9pt}c@{\hskip9pt}}
\toprule
Model  & Method & 
\small{ISIC2016} & 
\small{DFUC2020} & 
\small{BCCD} & 
\small{CPM-17} & 
\small{TBX11k} & 
\small{Luna16} & 
\small{ADNI} & 
\small{TN3k} &
\small{Avg} \\ \midrule
\multirow{8}{*}{GLIP-T} & Zero-shot            & 24.2          & 4.6          & 3.7          & 0.1          & 0.0          & 0.0          & 0.0          & 5.5          & 4.8 \\
                        & Linear Probing    & 38.8          & 30.3          & 48.2          & 29.9          & 8.3          & 5.7          & 7.6          & 26.3          & 24.4 \\
                        & Textual Inversion  & 49.3          & 42.2          & 56.7          & 20.3          & 6.5          & 25.8          & 13.5          & 51.6          & 33.2 \\
                        & Prompt Tuning     & \textbf{61.6 }         & 46.0          & 60.3          & \textbf{41.2 }         & 27.6          & 29.2          & 37.3          & 51.2          & 44.3 \\
                        & ConES (ours)       & 58.8          & \textbf{49.7 }         & \textbf{62.4}          & 40.7         & \textbf{33.4}          & \textbf{32.1 }         & \textbf{42.9}          & \textbf{56.6 }         & \textbf{47.1} \\
\cdashline{2-11}
                        & Full Model      & 59.6          & 50.1          & 62.9          & 44.0          & \textbf{41.5} & 40.9          & \textbf{48.9} & 58.4          & 50.8 \\

                        & ConES (ours)   & \textbf{63.1}          & \textbf{51.7} & \textbf{63.0} & \textbf{44.3} & 40.4          & \textbf{41.2} & \textbf{48.9} & \textbf{62.6}          & \textbf{51.9} \\ \midrule
\multirow{8}{*}{GLIP-L} & Zero-shot        & 24.6          & 3.6          & 10.6          & 9.9          & 0.1          & 0.0          & 0.0          & 8.1          & 7.1 \\
                        & Linear Probing    & 43.1          & 29.4          & 50.0          & 30.8          & 8.1          & 9.1          & 4.4          & 29.5          & 25.6 \\
                        & Textual Inversion & 44.2          & 47.5          & 58.6          & 32.0          & 13.5          & 18.7          & 20.7          & 47.5          & 35.3 \\
                        & Prompt Tuning     & \textbf{57.2 }         & 47.8          & 61.3          & \textbf{42.7}          & 22.8          & 25.9          & 31.9          & 48.5          & 42.3 \\
                        & ConES (ours)         & 56.6          & \textbf{51.1}          & \textbf{62.2}          & 40.4          & \textbf{26.5  }        & \textbf{29.0}          & \textbf{37.7}          & \textbf{56.9}          & \textbf{45.1 }\\
\cdashline{2-11}
                        & Full Model    & 66.8          & 51.5          & 63.4          & 44.7          & 37.9          & 41.4          & \textbf{48.4} & 61.3          & 51.9 \\
                        & ConES (ours)   & \textbf{67.0} & \textbf{53.1} & \textbf{63.5} & \textbf{44.8} & \textbf{40.5} & \textbf{43.6} & 48.2          & \textbf{64.3} & \textbf{53.1} \\
\bottomrule
\end{tabular}
}
\end{center}
\end{table*}